\pdfoutput=1
\documentclass[11pt]{article}
\usepackage[final]{acl}
\usepackage{adjustbox}
\usepackage{multirow}
\usepackage{multicol}
\usepackage{wrapfig}
\usepackage{scalerel}
\usepackage{float}

\usepackage[T1]{fontenc}    

\usepackage[utf8]{inputenc}

\usepackage{microtype}
\usepackage{svg}

\usepackage{inconsolata}

\usepackage[normalem]{ulem} 
\usepackage{twemojis}
\usepackage{soul}
\usepackage{times}
\usepackage{latexsym}
\usepackage{datetime}       
\usepackage{booktabs}       
\usepackage{nicefrac}       
\usepackage{xcolor}         
\usepackage{url}            
\usepackage{fnpct}           
\usepackage{enumitem}        
\usepackage{tcolorbox}       
\usepackage{csquotes}        
\usepackage{xspace}          
\usepackage{graphicx}
\usepackage{subfigure}
\makeatletter
\xspaceaddexceptions{\csq@qclose@i}
\makeatletter

\usepackage[scaled=0.85]{helvet}

\usepackage{bbm}
\usepackage{amsfonts}
\usepackage{amssymb}
\usepackage{mathtools}
\usepackage{amsmath}
\usepackage{bm}
\usepackage{amsthm}
\usepackage{thmtools} 
\usepackage{thm-restate}
\usepackage{cancel}

\usepackage{nicefrac}

\usepackage[createShortEnv,commandRef=Cref]{proof-at-the-end}

\newtheoremstyle{resultstyle} 
  {.8em} 
  {.8em} 
  {} 
  {} 
  {\bfseries} 
  {.} 
  {.5em} 
  {\thmname{#1 }\thmnumber{#2. }\textbf{\thmnote{#3}}}

\theoremstyle{resultstyle}

\usepackage{algorithm}
\usepackage{algpseudocode}  
\makeatletter
\algnewcommand{\LineComment}[1]{\Statex \hskip \ALG@thistlm \textcolor{blue}{// #1}}
\algnewcommand{\FirstLineComment}[1]{\Statex \hskip\ALG@tlm \textcolor{blue}{\(\triangleright\) #1}}
\algnewcommand{\InlineComment}[1]{\hfill\textcolor{blue}{\(\triangleright\) #1}}
\makeatother

\usepackage{cleveref}
\crefname{section}{\S}{\S\S}
\Crefname{section}{\S}{\S\S}
\crefformat{section}{\S#2#1#3}
\crefname{figure}{Fig.}{Fig.}
\crefname{alg}{Alg.}{Alg.}
\crefname{line}{line}{lines}
\crefname{appendix}{App.}{App.}
\crefname{equation}{eq.}{eqs.}
\crefname{table}{Table}{Tables}
\crefname{proposition}{Proposition}{Propositions}
\crefname{assumption}{Assump.}{Assumps.}
\crefname{lemma}{Lemma}{Lemmas}
\crefname{definition}{Defn.}{Defns.}
\crefname{hypothesis}{Hypothesis}{Hypotheses}
\crefname{estimator}{Estimator}{Estimators}
\crefname{theorem}{Theorem}{Theorems}
\crefname{thm}{Theorem}{Theorems}
\crefname{result}{Result}{Results}

\usepackage{tabularray}
\usepackage{cellspace}
\newcommand\cincludegraphics[2][]{\raisebox{-0.4\height}{\includegraphics[#1]{#2}}}
\usepackage{setspace}

\usepackage{siunitx}
\DeclareSIUnit[quantity-product = {}, reset-math-version = false]\thousand{k}
\DeclareSIUnit[quantity-product = {}, reset-math-version = false]\million{M}
\DeclareSIUnit[quantity-product = {}, reset-math-version = false]\billion{B}
\DeclareSIUnit[quantity-product = {}, reset-math-version = false]\trillion{T}

\DeclareSIUnit[quantity-product = {}, reset-math-version = false]\x{x}
\DeclareSIUnit[quantity-product = {}, reset-math-version = false]\percent{\%}

\DeclareSIUnit[quantity-product = {}, reset-math-version = false]\hour{h}
\DeclareSIUnit[quantity-product = {}, reset-math-version = false]\min{m}
\DeclareSIUnit[quantity-product = {}, reset-math-version = false]\sec{s}

\xspaceaddexceptions{\textsubscript}

\makeatletter
\DeclareRobustCommand*{\escapeus}[1]{%
  \begingroup\@activeus\scantokens{#1 }\endgroup}
\begingroup\lccode`\~=`\_\relax
   \lowercase{\endgroup\def\@activeus{\catcode`\_=\active \let~\_}}
\makeatother

\newcommand{\myemph}[1]{\textsf{{\escapeus{#1}}}}

\usepackage{amsmath}
\usepackage{xcolor}
\usepackage{tcolorbox}
\usepackage{tikz} 
\usetikzlibrary{shadings} 

\usepackage{twemojis}

\newif\ifshowemoji
\showemojitrue   

\newcommand{\bemoj}[1]{\ifshowemoji\texttwemoji{#1}\fi}

\title{What is the Best Sequence Length for \textsc{BabyLM}?}

\author{
    \textbf{Suchir Salhan}\thanks{Equal contribution}\bemoj{lemon} \bemoj{tangerine}  \quad
    \textbf{Richard Diehl Martinez}\footnotemark[1] \bemoj{lemon}  \\
    \textbf{Z\'ebulon Goriely} \bemoj{lemon}  \quad
    \textbf{Paula Buttery}\bemoj{lemon} \bemoj{tangerine} \\
    \bemoj{lemon}~Department of Computer Science \& Technology, University of Cambridge, U.K. \\
    \bemoj{tangerine}~ALTA Institute, University of Cambridge, U.K. \\
    \texttt{\{sas245, rd654, zg258, pjb48\}@cam.ac.uk}
}

\begin{document}
\maketitle

\begin{abstract}

Transformer language models typically operate with a fixed-length context window, which has grown in step with large-scale pretraining datasets. In the BabyLM Challenge, however, many past submissions have defaulted to using much shorter sequence lengths. 

We examine the impact of sequence length on BabyLM pretraining, to answer the simple question: what sequence length should we be using when training Baby LMs? Using 100M-word training data and fixed compute budgets, we compare 125M-parameter Mamba and OPT models, finding that although longer is often better, the optimal length depends on both task and architecture. Shorter sequences are sufficient for grammatical generalization tasks whereas longer contexts benefit morphological analogical reasoning tasks. 
 
\end{abstract}

\noindent

\begin{tblr}{colspec = {Q[c,m]|X[l,m]}, stretch = 0}
    \cincludegraphics[width=1.35em, keepaspectratio]{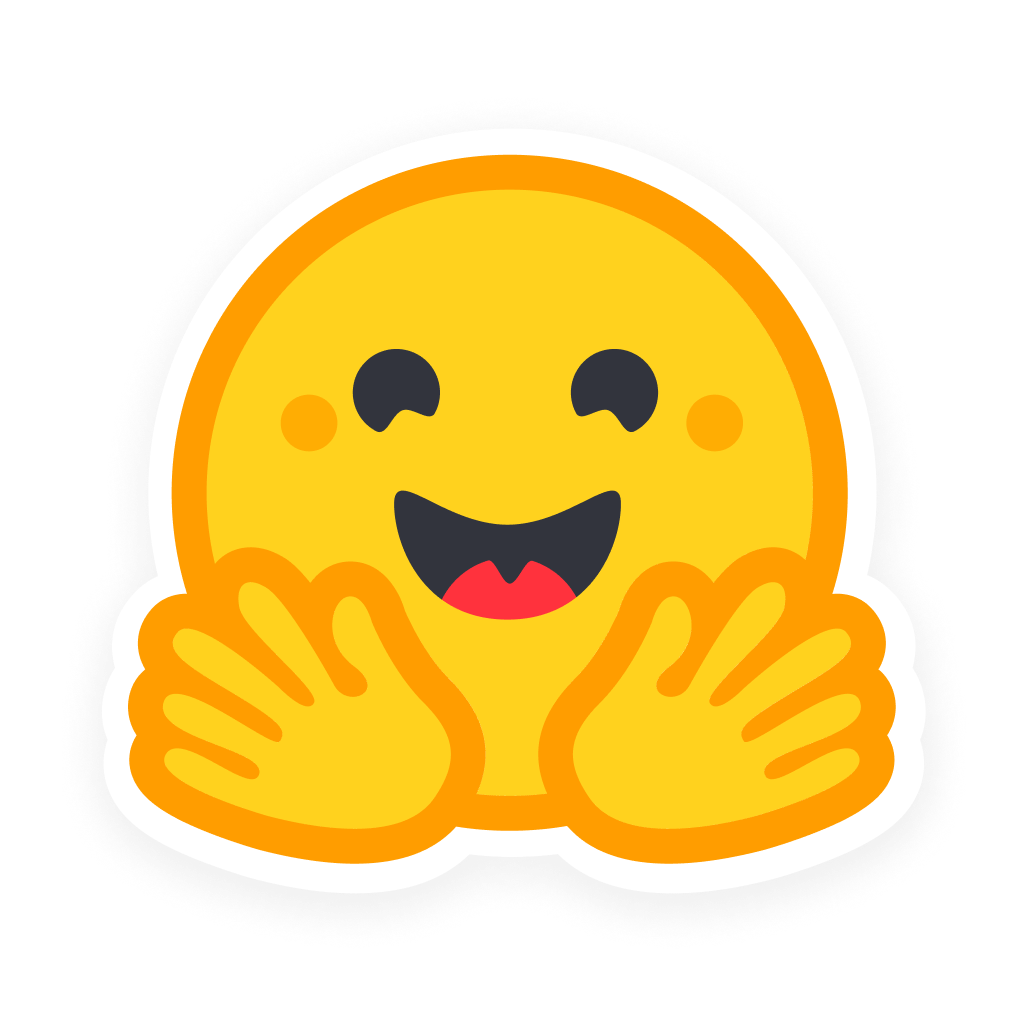} &
    \setstretch{.5}{\small\myemph{\textbf{How Long can You Go?} on \href{https://huggingface.co/babylm-seqlen}{HuggingFace} (models, tokenizers, and checkpoints)}} \\

    \cincludegraphics[width=1.1em, keepaspectratio]{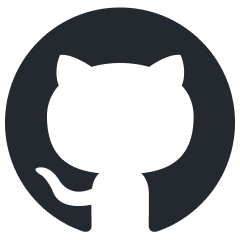} &
    \setstretch{.5}{\small\myemph{Training Code Open-Sourced on \href{https://github.com/rdiehlmartinez/babylm-seqlen}{GitHub}}}
\end{tblr}

\section{Introduction}

Transformer language models typically operate with a fixed context window, which has expanded in step with the growth of pre-training datasets --- from millions \citep{kiros2015skipthought} to trillions \citep{soldaini2024dolma} of tokens. Larger windows have improved performance on long-sequence reasoning tasks such as HellaSwag \citep{zellers2019hellaswag} and MMLU \citep{hendrycks2020mmlu}.

The BabyLM Challenge \citep{charpentier2025babylm} encourages researchers to revisit foundational assumptions in language-model pretraining. In this setting, models train on a 100M-token corpus, which may be repeated up to ten times for a total of 1B tokens. Under these  constraints, the belief that ``longer context is always better'' is less certain. Prior submissions to the challenge typically make use of shorter sequence lengths \citep{warstadt2023babylm}, often in an attempt to avoid training instability given the restricted data and as a cognitively-inspired attempt to mimic human working memory limitations \citep{cheng2023mcbabylm}.

Our starting question is simple: what happens if we train a BabyLM using the same methods typically applied at large scale? Many submissions implicitly assume that small batch sizes and short sequences are both cognitively plausible and optimal under limited data. But is this actually true?

\paragraph{The Case for Long Sequences}
The main benefit of training language models with longer sequence lengths is \textbf{training efficiency}. Longer sequence lengths allow the model to observe more tokens per step, provide more learning signal per update, and reduce the noise in gradient estimates. 

\paragraph{The Case for Small Sequences} However, in the data constrained setting of the BabyLM challenge, using larger sequences means models are updated less often; smaller sequences, despite yielding noisier gradient approximations, enable models to be updated more overall. 

\vspace{1em}

These trade-offs motivate our first research question: \textbf{what is the optimal sequence length for each BabyLM evaluation task?} We explore optimality both in terms of the sequence length that produces the highest score at the end of training, as well as a more nuanced analysis that considers training time. 

Next, we explore a second related question: \textbf{does this optimal length depend on the model \emph{architecture}?} State Space Models (SSMs) are particularly interesting here: by removing the $n^2$ state-storage requirement of self-attention, they may handle long sequences more efficiently than Transformers.

To investigate, we train two BabyLM families—one using the Open Pre-Trained Transformer (OPT) \citep{zhang2022opt}, the other using Mamba \citep{gumamba}—on the 100M \textsc{Strict} BabyLM dataset, varying only input sequence length. We span short contexts (64 tokens) common in cognitively-inspired setups to very long contexts (8192 tokens) typical in modern LLMs. 

Results show that the ideal sequence length for training language models depends heavily on both the specific task and the model architecture. For some tasks, such as syntactic evaluation benchmarks, shorter sequences provide better performance and faster training times. In contrast, tasks that require understanding longer context, like entity tracking or reading comprehension, benefit from much longer sequences, sometimes up to 8192 tokens. When comparing model architectures, we find that the OPT Transformer generally performs best with a wider range of sequence lengths, including very long contexts, while the Mamba state-space model tends to achieve near-optimal results using shorter or moderate-length sequences. This suggests that different sequence-length strategies may be needed depending on the model's design and the nature of the task. We provide a set of sequence length recommendations for BabyLM practioners aiming to balance training efficiency and model performance. Selecting a training sequence length tailored to the specific task and model architecture can significantly reduce computational costs and training time without sacrificing accuracy, with the added benefit of making pretraining BabyLMs more accessible and environmentally friendly.

\section{Background}


\subsection{Sequence Length and Modern Language Models}

Multiple studies suggest that shorter sequence lengths can benefit smaller language models, particularly under data constraints. In the BabyLM setting, \citet{cheng2023mcbabylm} report that using individual sentences and avoiding sequence packing yields better results, with sequences as short as 32 tokens outperforming 512-token contexts. \citet{warstadt2023babylm} similarly note that many top submissions to BabyLM used short contexts, aligning with developmental-learning constraints and maximizing limited data efficiency.

Outside BabyLM, compute-efficient training approaches also favor short sequences. Both \citet{izsak2021bertbudget} and the original BERT work \citep{devlin-etal-2019-bert} train primarily with 128-token sequences before a final phase at 512 tokens, while \citet{geiping2023cramming} find 128 tokens sufficient for strong downstream performance even with larger datasets. The LTG-BERT model from the first BabyLM Challenge adopts the same 128-to-512 token schedule \citep{samuel2023ltgbert}.

\subsection{Sequence Length Across Architectures: Transformers and State-Space Models}
\label{background-architecture}

Sequence length \(L\) plays different roles across architectures. In Transformers, \(L\) defines a fixed input window for both training and inference, directly determining attention cost. Inputs longer than the maximum \(L\) must be truncated or handled with long-context extensions such as structured attention \citep{hao2022structured} or compression \citep{li2023compressing}. Length extrapolation methods adjust positional embeddings to process sequences beyond the trained \(L\) \citep{press2021train, chen2021simple, su2024roformer}, while interpolation integrates new information into existing positions \citep{chen2023extending}.

By contrast, recurrent models and State Space Models (SSMs) such as Mamba do not impose a hard cap on \(L\). Mamba retains memory via parameterized state-space dynamics, capturing long-range dependencies with linear scaling \citep{gumamba}. Trained with sequences up to \(L = 2048\), it can carry compressed history across chunks, making long contexts less costly in memory and computation. These differences suggest that Mamba may have a higher training-optimal \(L\) than a vanilla Transformer like OPT, owing to its more efficient handling of long-range information.

 \subsection{Sequence Length, Working Memory, and Psychometric Plausibility}

The use of shorter sequence lengths aligns with findings in cognitive modeling. A central idea in Cognitive Science is that working-memory limitations can, paradoxically, aid language learning by imposing a recency bias and promoting abstraction through chunking \citep{newport1988constraints, christiansen2016now, wilcox2025bigger}.  

\citet{elman1990finding} showed that recurrent neural networks trained on simple, short sequences in early learning stages were better at acquiring syntactic generalizations. This ``starting small'' strategy reflects two hypotheses: (i) learners may benefit from gradually increasing input complexity rather than starting with long or complex sequences \citep{bengio2009curriculum}, a principle used in Curriculum Learning approaches to the BabyLM Challenge \citep{martinez-etal-2023-climb, salhan-etal-2024-less}; and (ii) memory limitations act as a resource constraint, forcing language input to be ``chunked'' into storable, manipulable units.  
This second view has motivated BabyLM approaches that incorporate cognitively inspired working-memory constraints \citep{armeni2022characterizing, mita2025developmentally, de-varda-marelli-2024-locally, thamma2025human, clark2025linear}. For example, \citet{thoma-etal-2023-cogmemlm} adopt a maximum sequence length of 512 for their CogMemLM architecture.

\section{Methodology}

We train OPT and Mamba models on the \textsc{Strict} 100M subset of the \textsc{BabyLM} corpus \citep{charpentier2025babylm} using sequence lengths ranging from 64 to 8192 tokens. Our goal is to identify a sequence length \(L^*\) that balances task performance with computational efficiency.

\subsection{Default Model Hyperparameters}
\begin{table}[h!]
\centering
\begin{tabular}{lcc}
\toprule
\textbf{Parameter} & \textbf{Mamba} & \textbf{OPT} \\
\midrule
\texttt{vocab\_size} & 50257 & 50272 \\
\texttt{hidden\_size} & 768 & 768 \\
\texttt{num\_hidden\_layers} & 32 & 12 \\
\texttt{state\_size} & 16 & -- \\
\texttt{expand / ffn\_dim } &  2 & 3072 \\
\texttt{num\_attention\_heads} & -- & 12 \\
\texttt{hidden\_act} & silu & relu \\
\bottomrule
\end{tabular}
\caption{Key default hyperparameters for \texttt{MambaConfig} and \texttt{OPTConfig} as implemented in Hugging Face Transformers.}
\label{tab:key-defaults}
\end{table}
We include a full table of training hyperparameters in \textit{Table} \ref{tab:hyperparams_main}. 

\subsection{Model Families}
We train two model families: one based on the OPT architecture and the other on Mamba. A custom tokenizer is trained on the full BabyLM training set, starting from the Byte-Pair Encoding (BPE)-based GPT-2 tokenizer provided by Hugging Face \citep{sennrich2016bpe}, then retrained on the BabyLM dataset.  For each model family, we train models with and without warmup. In our warmup models, we scale the learning rate linearly with sequence length, using 64 tokens as a reference, to maintain approximately constant per-token updates across sequences from 128 to 4096 tokens, and increase it gradually from zero during a warmup period to stabilize early training. We follow the checkpointing logic required for submission models in the 2025 Shared Task \citep{charpentier2025babylm}, saving checkpoints at increasingly intervals. 

\subsection{Dataset Preparation}
The \textsc{BabyLM} training corpus is shuffled at the document level, tokenized, and split into fixed-length chunks matching the target sequence lengths: 64, 128, 256, 512, 1024, 2048, 4096, and 8192 tokens. This produces eight distinct datasets, one for each sequence length. Key hyperparameters for the two model configurations are listed in \cref{tab:key-defaults} and we open-source our trained models and the eight prepared datasets.\footnote{\emph{url anonymized for review.}} 

\subsection{Training-Optimal Sequence Length}
Our setup allows us to examine the trade-off between sequence length, task performance, and computational cost in a controlled manner. Let \(M(L)\) denote a BabyLM model trained with sequence length \(L\), and \(E\) a BabyLM evaluation task. If two models \(M(L_1)\) and \(M(L_2)\) achieve comparable accuracy on \(E\), but \(T(M(L_1)) \ll T(M(L_2))\) in training time, we consider \(M(L_1)\) the more \emph{training-optimal} choice for \(E\).  

We define the \textbf{training-optimal sequence length} \(L^*\) for task \(E\) as the shortest \(L\) that yields competitive accuracy relative to other lengths while offering a measurable training-time benefit. Training time is expressed as a proportion of the longest run within the same model family to facilitate comparison under setup variance and without exhaustive hyperparameter sweeps.

\subsection{Evaluation}
We report \(L^*\) for each model family (OPT and Mamba) and each evaluation task in the BabyLM Evaluation Pipeline. This addresses two research questions:

\begin{enumerate}
    \item \textbf{Task-level trends:} Do values of \(L^*\) show consistent patterns across BabyLM evaluation tasks \(E\)?
    \item \textbf{Architecture-level trends:} Do differences in \(L^*\) between Mamba and OPT reflect their distinct sequence-handling mechanisms, as discussed in Section~\ref{background-architecture}?
\end{enumerate}

While a single \(L^*\) that improves performance across all tasks is unlikely, some practitioners may wish to optimize for overall leader board performance (e.g., maximizing the ``text-average'' score across zero-shot tasks), whereas others may target specific benchmarks such as BLiMP \citep{warstadt-etal-2020-blimp-benchmark} or \textit{psychometric fit}. The latter, introduced in the 2025 Shared Task \citep{charpentier2025babylm}, comprises two tasks:  

\begin{itemize}
    \item \textbf{Wug Adjectival Nominalisation} \citep{hofmann2025derivational} — tests morphological analogical generalisation, e.g., \textsc{available} \(\to\) \textsc{availability}.
    \item \textbf{Readability Prediction} \citep{de2024cloze} — evaluates model alignment with human processing by correlating cloze probabilities with human predictability ratings from self-paced reading and eye-tracking data.
\end{itemize}

\section{Results}

\subsection{Optimal Sequence Length, \(L*\), for BabyLM Evaluation Task}

\begin{figure*}
    \centering
    \includegraphics[width=\linewidth]{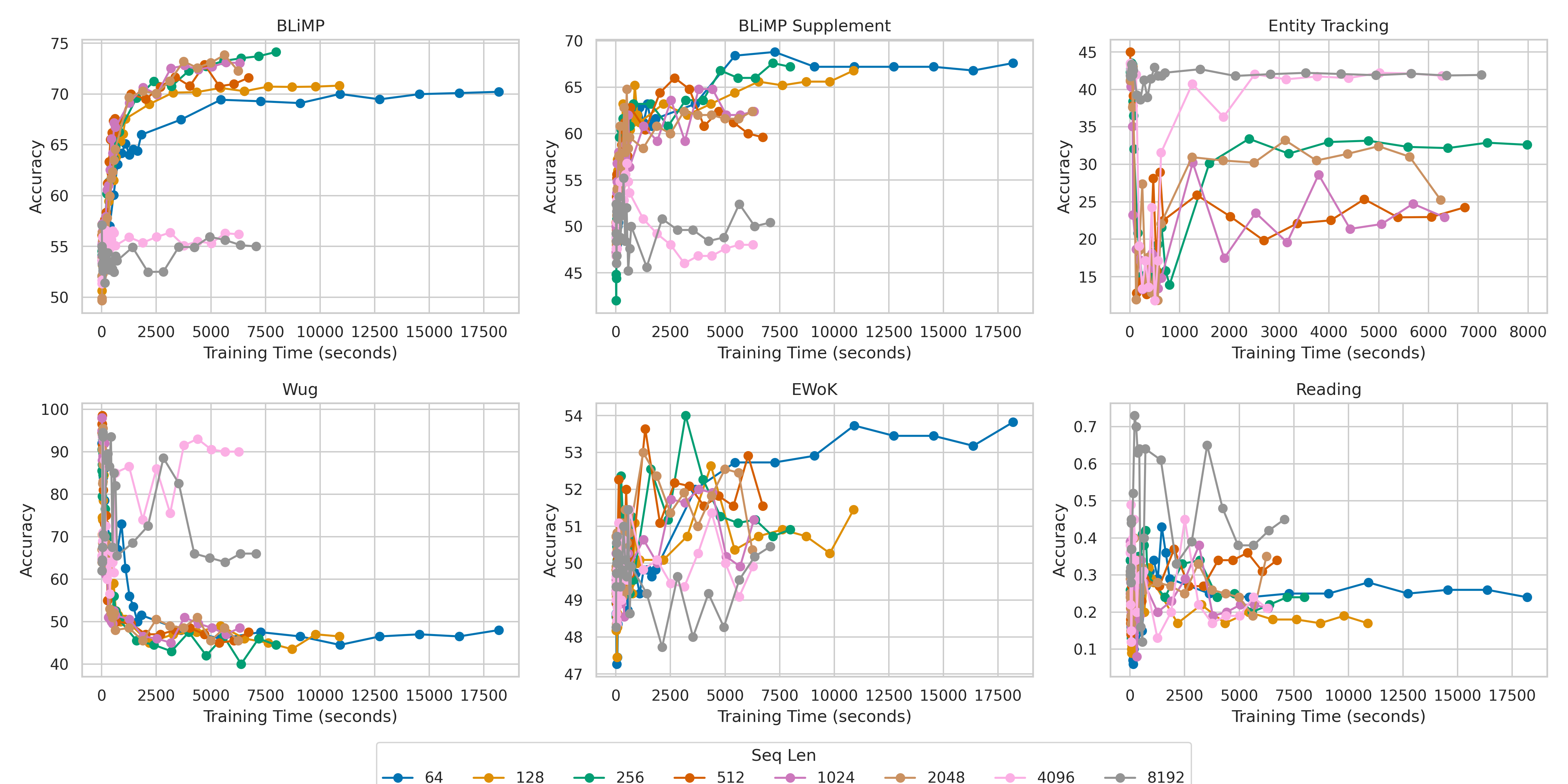}
    \caption{OPT Model Families: Effect of Sequence Length Accuracy vs Training Time per Metric. Evaluation of OPT 125M Family trained on 100M \textsc{Strict} BabyLM Corpus with Warmup with a range of sequence lengths \(\{64, 128, 256, 512, 1024, 2048, 8192 \}\) on the Zero-Shot Evaluation Tasks of the 2025 BabyLM Evaluation Pipeline \citep{charpentier2025babylm}}
    \label{fig:opt-training-time}
\end{figure*}

In \textit{Figure} \ref{fig:opt-training-time}, we plot the training time for OPT model with different sequence lengths. This shows accuracy of eight OPT 125M parameter models trained on the 100M \textsc{Strict} corpus across training, plotted against the training time for each model. The figure only shows results for the OPT family with warmup (see \textit{Table} \ref{tab:mamba_opt_results_split} for full results). Using the training time data, we can identify the \emph{training-optimal sequence length from the OPT model family} \(L_{OPT}^*\) for each BabyLM evaluation task by selecting the shortest sequence length that still achieves near-peak performance. 

\textbf{The effect of sequence length is task-dependent across BabyLM Evaluation Tasks.} We find that the effect of sequence length is inconsistent across tasks in the 2025 BabyLM Evaluation Pipeline \citep{charpentier2025babylm}.   There is a non-monotonic benefit of sequence length.

\textbf{General Trends.} Shorter sequence lengths perform better on \textbf{BLiMP} and \textbf{BLiMP Supplement}. The best performance on BLiMP is obtained by our \texttt{opt-256} model, while \texttt{opt-64, opt-128} and \texttt{opt-256} obtain similar performance on BLiMP Supplement, with performance generally declining as sequence length increases beyond 1024 tokens.

Our shortest sequence length model \texttt{opt-64} obtains the highest accuracy on the \textbf{EWoK} benchmark, however, it remains largely stable across sequence lengths, suggesting that EWoK tasks are less sensitive to the sequence length. 

Conversely, longer sequence lengths perform better on Entity Tracking, Wug and Reading Evaluation Tasks. We can an opposite pattern for BLiMP and BLiMP Supplement. For OPT, \textbf{Entity Tracking} performance shows modest sensitivity to sequence length, with no consistent upward trend as sequence length increases. While mid-range sequences (256–1024 tokens) achieve comparable scores, extreme lengths (4096–8192 tokens) exhibit more variable results, indicating that longer contexts do not reliably improve entity-tracking capabilities. However, shorter sequence length models generally perform poorly on the Entity Tracking task, with \texttt{opt-256} achieving an accuracy of \(32.42 \%\). 

For OPT,  performance on the \textbf{Wug} evaluation task strongly benefits from longer sequence lengths, particularly at 4096–8192 tokens with warmup, where accuracy reaches up to \(90 \%\). This suggests that \textbf{longer sequence lengths might support learning productive morphological patterns and generalizing to novel forms}.

Overall, these results indicate that OPT’s optimal sequence length is highly task-dependent: shorter sequences support better BLiMP performance, whereas longer sequences support lexical productivity tasks, like Wug, and Entity Tracking.


\begin{table*}[h!]
\setlength{\tabcolsep}{4pt}
\centering
\begin{tabular}{|l c c | c c || c c | c c|}
\hline
\textbf{Task} & \multicolumn{4}{c||}{\textbf{OPT}} & \multicolumn{4}{c|}{\textbf{Mamba}} \\ \hline
 & \textbf{$L^*$} & \% (Longest) & $L_{\text{best}}$ & \% (Longest) & \textbf{$L^*$} & \% (Longest) & $L_{\text{best}}$ & \% (Longest) \\
\hline
BLiMP & 1024 & 34.8 & 64   & 100.0 & 512  & 37.3 & 2048 & 33.3 \\
BLiMP Suppl. & 256  & 43.9 & 64   & 100.0 & 64   & 100.0 & 64   & 100.0 \\
Entity Tracking & 4096 & 34.5 & 8192 & 38.8 & 1024 & 35.2 & 128  & 58.4 \\
Wug & 4096 & 34.5 & 4096 & 34.5 & 128  & 58.4 & 128  & 58.4 \\
EWoK & 4096 & 34.5 & 2048 & 34.3 & 1024 & 35.2 & 512  & 37.3 \\
Reading & 8192 & 38.8 & 8192 & 38.8 & 512  & 37.3 & 64 & 100 \\
\hline
\end{tabular}
\caption{Training-optimal sequence lengths $L^*$ and best-performing lengths $L_{\text{best}}$ for OPT and Mamba models on BabyLM evaluation tasks, with training time as a percentage of the longest training time for that model.}
\label{tab:optimal-seq}
\end{table*}

\subsection{Model Architecture: Mamba and OPT}

\begin{figure*}
    \centering
    \includegraphics[width=\linewidth]{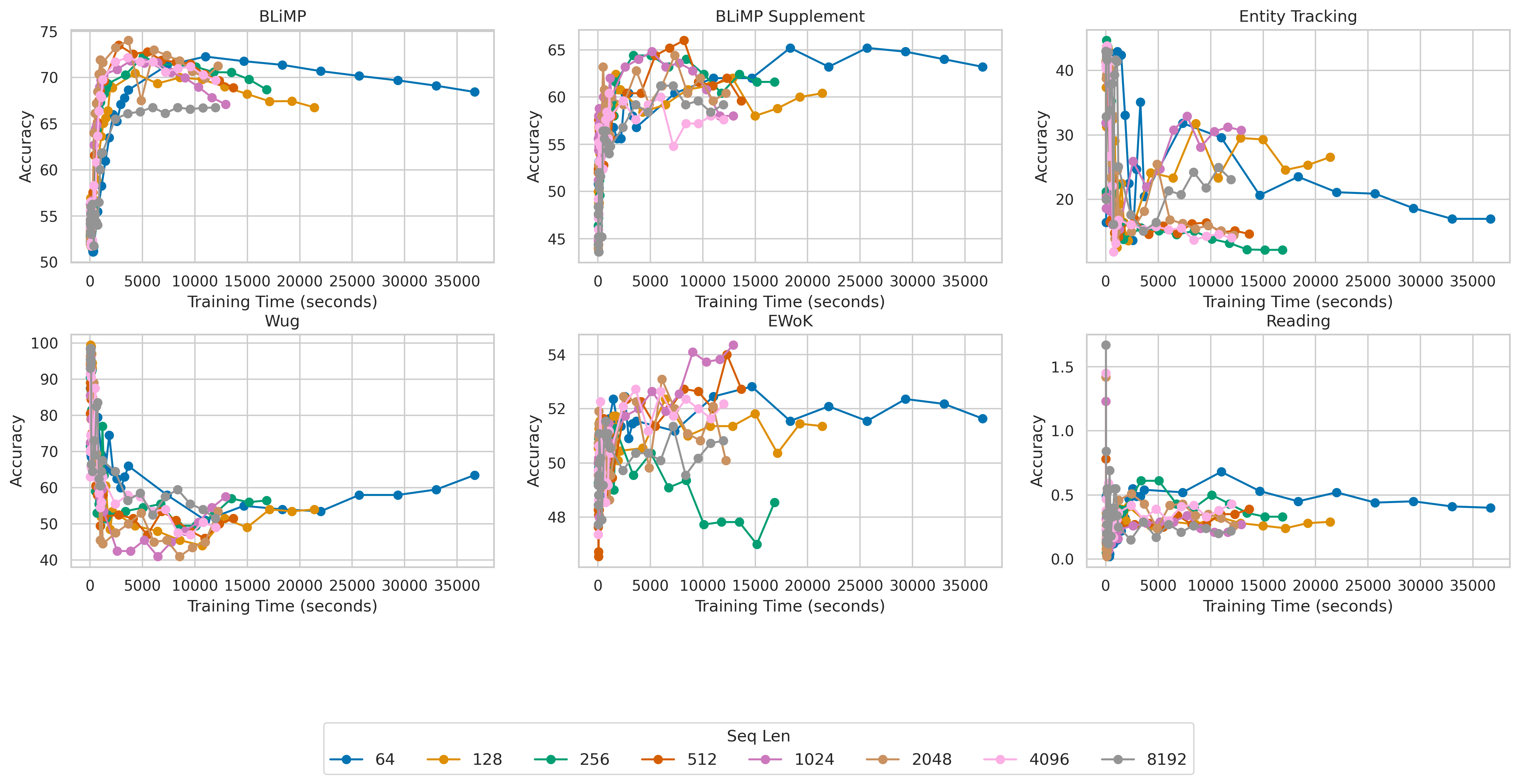}
    \caption{Mamba Model Families: Effect of Sequence Length Accuracy vs Training Time per Metric }
    \label{fig:mamba-training-time}
\end{figure*}

We similarly report \( L_{\text{Mamba}}^* \) for each BabyLM evaluation task. Scaled training time–accuracy curves for our Mamba Family are shown in \textit{Figure}~\ref{fig:mamba-training-time}. Table~\ref{tab:optimal-seq} shows the training‐optimal sequence lengths (\( L^* \)) and the lengths yielding the best evaluation performance (\( L_{\text{best}} \)) for OPT and Mamba across BabyLM tasks, alongside training cost relative to the longest‐context setting.

Mamba achieves slightly lower performance than OPT across most benchmarks, often matching or slightly exceeding OPT on mid-range context tasks, while OPT tends to dominate in long-context tasks. For instance, on BLiMP and BLiMP Supplement, Mamba reaches comparable scores to OPT despite shorter sequence lengths, but in general, performance is lower than OPT. On Entity Tracking, a long-range dependency task, Mamba performs best at sequence lengths of 128–1024 tokens, whereas OPT benefits from much longer contexts (up to 8192 tokens). However, again, performance is generally lower than OPT. On Wug and EWoK, Mamba generally performs comparably to OPT at moderate sequence lengths (128–512 tokens). On Wug, Mamba outperforms OPT on nearly all sequence lengths, except the longest sequence lengths (4096). Mamba's EWoK performance is comparable to OPT but consistently obtains a marginally lower accuracy. We include a full table of results (\textit{Table} \ref{tab:mamba_opt_results_split}) that provides a side-by-side comparison of Mamba and OPT. 

The Reading results exhibit a striking pattern: Mamba achieves its peak score using the shortest context (64 tokens), whereas OPT continues to improve up to 8192 tokens. These results highlight task-specific differences in optimal context requirements between the two model families. Examining sequence length optimality, we observe that Mamba consistently prefers mid-range sequences ($L$ between 64 and 1024 tokens) for training efficiency and evaluation performance, whereas OPT exhibits a wider spread ($L$ between 256 and 8192 tokens). 

Comparing Learning Dynamics, Mamba often attains near-peak evaluation performance with substantially shorter sequences than OPT, implying faster training times and reduced computational cost without substantial loss in accuracy. This behavior suggests that the Mamba architecture effectively leverages its hybrid attention mechanisms to capture both local and moderately long-range dependencies, reducing the necessity for extremely long contexts that OPT requires for certain tasks.

Table~\ref{tab:optimal-seq} offers actionable guidance for selecting efficient sequence lengths. For practitioners using \textbf{OPT}, we recommend $L^* = 256$ or $512$ for syntax-sensitive tasks like BLiMP and BLiMP Supplement, achieving 35--44\% of the full training cost while retaining high accuracy. For tasks requiring long-range dependencies, such as Wug, Entity Tracking, and Reading, longer contexts ($L^* = 4096$ or $8192$) yield meaningful gains but at higher computational cost. Practitioners can adopt $L = 2048$ as a reasonable default for OPT to balance efficiency and generality across BabyLM tasks.

For \textbf{Mamba}, $L^*$ values tend to cluster at shorter lengths. We recommend $L = 64$ or $128$ for BLiMP Supplement, Wug, and Reading, where training time can be reduced by up to 60--65\% without significant accuracy loss. Mamba’s performance on EWoK and Entity Tracking is best at mid-range lengths ($L = 512$–$1024$), suggesting practitioners should avoid unnecessarily long contexts for most tasks. Overall, $L = 512$ offers a safe and efficient baseline across both architectures when training budget or time is limited. These recommendations allow users to reduce compute overhead while maintaining competitive task-level performance.

\subsection{Psychometric Plausibility and Sequence Lengths} \label{psych-plausibility}

\begin{figure}
    \centering
    \includegraphics[width=\linewidth]{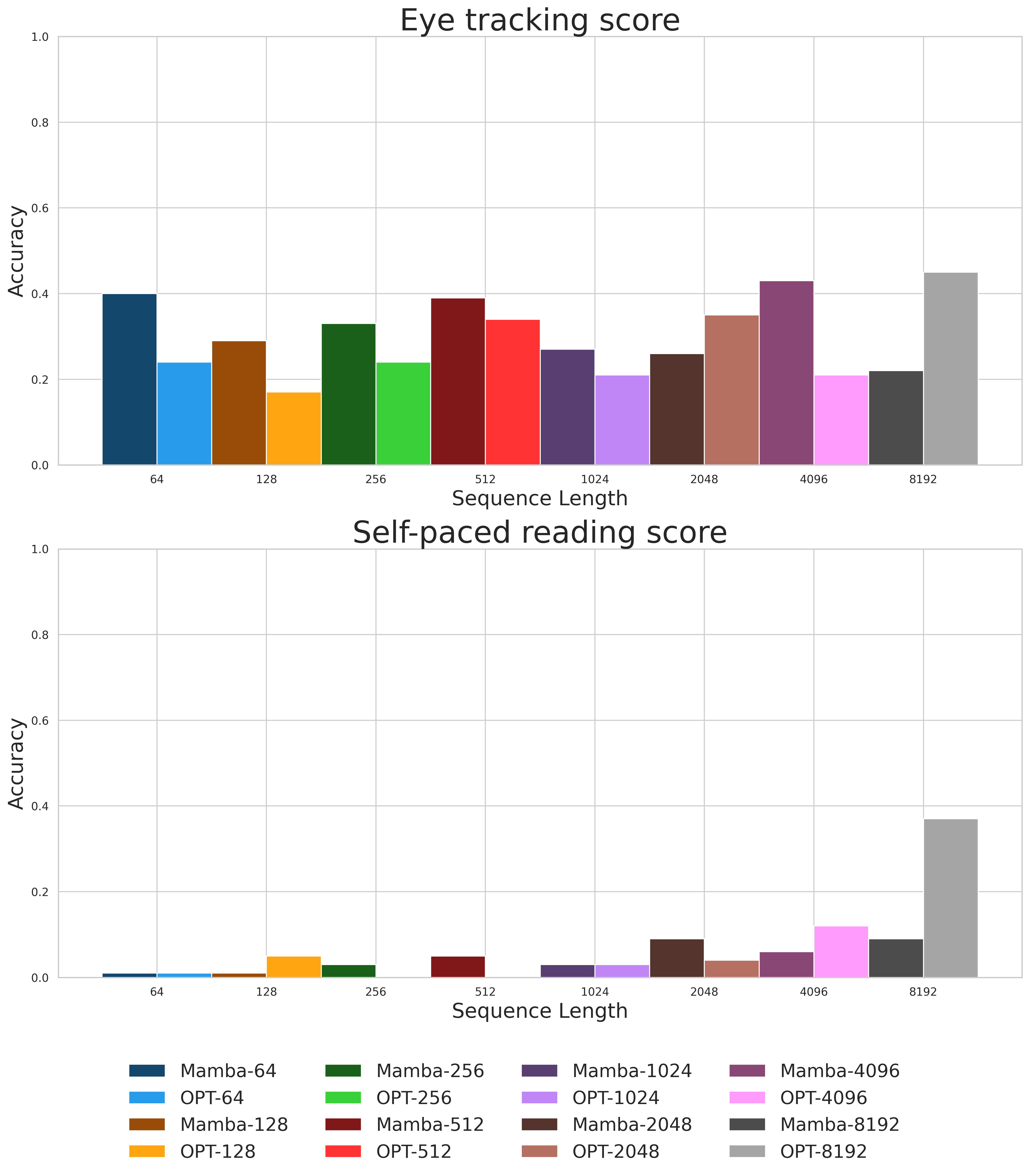}
    \caption{Distribution of Reading Sequence Length Model Accuracies for OPT Architecture }
    \label{fig:psychometric-fit}
\end{figure}

\textit{Figure} \ref{fig:psychometric-fit} reports the evaluation of the OPT family on the readability prediction task \citep{de-varda-marelli-2024-locally}. 

We evaluate model performance on two psycholinguistic benchmarks—eye-tracking and self-paced reading—across varying input sequence lengths. As shown in Figure~\ref{fig:psychometric-fit} (top), Mamba models exhibit relatively stable eye-tracking scores as context length increases, consistently outperforming their OPT counterparts at longer contexts (e.g., Mamba-4096 vs.\ OPT-4096). Notably, OPT-8192 achieves the highest accuracy ($\sim$0.45), indicating improved alignment with human eye-tracking behavior for extended inputs. In contrast, OPT models show more variable performance, with a decline in accuracy at mid-to-long sequence lengths, followed by a modest recovery at 8192 tokens.

For the self-paced reading benchmark (Figure~\ref{fig:psychometric-fit}, bottom), accuracy is generally lower across both model families, reflecting the greater challenge of modeling human reading times. Only the OPT-8192 configuration achieves a notable gain ($\sim$0.35), suggesting that long-context processing is critical for capturing self-paced reading patterns. While Mamba models outperform OPT at intermediate lengths (e.g., Mamba-2048 vs.\ OPT-2048), they fall short at the longest context window, indicating potential limitations in modeling long-range syntactic and semantic dependencies effectively.

Overall, Mamba outperforms OPT on eye-tracking prediction at long contexts, suggesting some alignment with incremental human sentence processing. However, OPT recovers and exceeds Mamba on self-paced reading at very long contexts.

\section{Discussion}

Our results suggest that sequence length plays a central, task-sensitive role in small-scale language modeling, particularly within the BabyLM benchmark suite. Rather than observing a monotonic relationship between longer sequences and better performance, we find that each task exhibits a distinct profile of sequence length sensitivity. This challenges the default practice of adopting a single sequence length for all training and evaluation scenarios and suggests that per-task tuning of input length may yield significant efficiency gains without sacrificing accuracy.

\subsection{Effect of Model Architecture}

When comparing architectures, we find that \textbf{OPT and Mamba differ substantially in their sequence length dynamics}. The OPT family benefits from long contexts on tasks like Reading and Wug, with optimal sequence lengths ($L^*$) extending up to 8192 tokens. In contrast, tasks such as BLiMP and EWoK reach peak or near-peak performance at much shorter lengths (64--256 tokens). 

This heterogeneity is likely task-related and reflects the diversity of  BabyLM tasks. As the evaluation pipeline incorporates more tasks, there are differences in the types of linguistic structures that they emphasise---e.g., syntactic locality in BLiMP versus document-level coherence in Reading---making sequence length a proxy for task-specific inductive biases. This makes it challenging to develop one model that performs uniformly well across all tasks.  Additionally, \textit{Figure} \ref{fig:opt-training-time} reveals pronounced fluctuations in training performance across sequence lengths, particularly for Wug and other productivity-oriented tasks. Many models show declining accuracy after initial progress, indicating that longer training does not always improve evaluation outcomes. For these tasks, shorter or mid-range sequences lengths achieve near-peak accuracy faster, reducing both computation and potential overfitting. From a practical perspective, compute-efficient training to improve performance on these tasks may involve early stopping after a moderate number of updates—around 512 steps in our experiments.

Compared to our expectations of the differences between Transformers and SSM-based architectures like Mamba, the observed OPT results only partially align. Unlike Transformers, which pay a quadratic compute and memory cost for longer contexts, and unlike RNNs, which must propagate hidden states step-by-step, Mamba’s recurrence-style state updates allow it to scale more gracefully with window size. While we predicted an optimal range of 256–1024 tokens for most tasks, some OPT tasks indeed peaked in this mid‐range, but others (notably Reading and Entity Tracking) favored much longer sequences than expected, suggesting certain BabyLM subtasks draw more heavily on full‐document context. For Mamba, the findings diverge more strongly from our forecast. We anticipated a right‐shifted optimum (1024–4096) and broad benefits from longer windows, yet $L^*$ clustered at shorter lengths (64–1024) and $L_{\text{best}}$ often appeared in the lower mid‐range. Despite Mamba’s architectural promise---continuous-time dynamics and implicit memory---we observe that \textbf{Mamba consistently prefers shorter or mid-range sequence lengths}, with $L^*$ clustering between 64 and 1024 tokens. While this allows Mamba to train more efficiently than OPT on average, it often lags in final performance, particularly on tasks requiring sustained access to long-range dependencies (e.g., Entity Tracking, Reading). These results complicate expectations from prior work \citep{gumamba} suggesting Mamba-like models can exploit long contexts more effectively than Transformers. In our small-model, low-data regime, Mamba’s theoretical capacity may be bottlenecked by optimization constraints or underutilized due to limited token diversity.

While Mamba’s theoretical strengths in long-context modeling are appealing, fully realizing these advantages may require larger models, more diverse data, or improved optimization strategies. Future work should systematically disentangle these factors to determine whether the observed limitations are fundamental to the architecture, a consequence of optimization dynamics, or an artifact of the data scale. The consistent preference of Mamba for shorter sequences raises important questions. One possibility is that this reflects an architectural limitation: despite Mamba’s theoretically continuous-time, state-space recurrence dynamics, the model may be unable to store and retrieve fine-grained information over very long sequences at small model scales. Another contributing factor could be optimization challenges: gradient diversity and update counts may be insufficient in the 100M-token regime to fully exploit long-range dependencies. Finally, data-scale constraints may limit Mamba’s capacity to generalize across long contexts, since small datasets provide fewer instances of extended dependency structures for learning.

These findings suggest that Mamba’s efficiency – achieving near-peak performance with shorter sequences – can reduce training time and computational cost, offering a practical advantage in low-resource or small-model settings. Nevertheless, this efficiency comes at a trade-off: for tasks where long-range dependencies are critical, OPT’s Transformer-based architecture remains superior, even at the expense of substantially higher training costs. This aligns with previous observations for RNN and SSM variants in small-data regimes \citep{haller-etal-2024-babyhgrn}, emphasizing that architectural efficiency does not automatically translate into performance gains in low-data or small-scale contexts.

In practical terms, our results offer guidance for model selection and sequence length configuration. For OPT, shorter sequences (256–512 tokens) suffice for syntax-sensitive tasks, while longer sequences (4096–8192) are beneficial for document-level and productivity tasks. For Mamba, mid-range sequences (128–512 tokens) generally balance performance and efficiency, though extreme long contexts rarely yield additional gains. When compute budgets are limited, using Mamba with shorter sequences may provide a favorable trade-off between training time and accuracy, while OPT remains the model of choice for tasks with high long-range dependency demands.

This suggests that, at small‐model scale and 100M word budgets, Mamba’s state‐space recurrence may not fully exploit very long contexts—possibly due to limited capacity to store fine‐grained long‐range information, or a stronger dependence on update count and gradient diversity than hypothesized.  This mismatch invites further scrutiny into how scaling laws and data regimes modulate sequence-length utility. The BabyLM setting---100M tokens and training using an architecture with 125M parameters---imposes strong bottlenecks on both parameter and data capacity. For Transformers like OPT, longer contexts may serve to increase gradient diversity and reinforce context-sensitive representations, whereas Mamba may compress or discard such information more aggressively. The result is a modest gain in training efficiency, but with diminished generalization on long-context benchmarks. These trade-offs are particularly relevant to BabyLM’s goal of modeling developmentally plausible language learning with limited resources.


\subsection{Sequence Length and Psychometric Plausibility}

From a cognitive perspective, our sequence length results provide direct computational support for the \textbf{“starting small” hypothesis}  \citep{elman1990finding, newport1988constraints}. In \textit{Section} \ref{psych-plausibility}, we observed that syntactic tasks like BLiMP consistently reach peak performance at shorter sequences (64–256 tokens), a pattern that suggests limiting context during learning can facilitate chunking, abstraction, and generalisation. This mirrors the cognitive insight that constrained working memory during early language exposure can promote more robust syntactic representations. Importantly, these findings are not merely incidental: they indicate that the empirical optima for sequence length in small-scale language models align with theoretically motivated cognitive constraints, showing that “starting small” can confer measurable learning advantages even in artificial systems.

Mamba’s recurrent, state-based architecture provides a compelling demonstration of this principle in practice. By maintaining local state updates and implicitly emphasizing recent context, Mamba performs stably at shorter sequences, despite having the capacity for longer-range memory. This alignment between architectural design and empirical sequence length optima suggests that Mamba operationalizes a cognitively inspired inductive bias: the model leverages local context efficiently to capture syntactic regularities, providing a computational analogue to human working memory limitations. In contrast, OPT benefits from long sequences primarily on tasks requiring document-level integration, such as Reading or Entity Tracking, highlighting how different architectures interact with sequence length in ways that parallel the cognitive distinction between local syntactic processing and global discourse comprehension.

The psycholinguistic benchmarks further reinforce this link. Mamba’s locally-informed processing produces smoother, word-by-word plausibility predictions, echoing human recency effects in reading, whereas OPT’s global attention facilitates retention and manipulation of hierarchical or discourse-level structures. This complementary pattern suggests that model architecture and sequence length interact to capture different aspects of linguistic cognition: recurrence-based models like Mamba naturally encode inductive biases favoring short, syntactically rich sequences, while attention-based Transformers excel when broader context is required.


For BabyLM practitioners, we hope our results provide a practical, resource-conscious strategy for selecting sequence lengths in low-resource language modeling. By computing the training-optimal length $L^*$ for each evaluation task $E$, practitioners can identify the shortest sequence that delivers near-peak performance at a fraction of the training cost. This allows for more efficient model development, particularly in constrained environments where compute or wall-clock time is limited. Rather than relying on fixed defaults (e.g., $L=512$ or $L=2048$), users can adopt our methodology to empirically select task-appropriate sequence lengths for their architecture of choice. As we demonstrate, $L^*$ often varies across tasks and model types, and even small adjustments can yield substantial training-time savings without sacrificing downstream accuracy.

Taken together, our findings suggest that \textbf{no single sequence length is optimal across tasks, models, or metrics}.  For BabyLM, this heterogeneity means leaderboard design and evaluation strategy should account for task-specific sequence length sensitivity. For example, syntactic tasks like BLiMP reach peak performance at short sequences (64–256 tokens), whereas discourse-heavy tasks like Reading or Entity Tracking benefit from much longer contexts (up to 8192 tokens). A practical approach would be to report, for each task, performance at the training-optimal sequence length (\(L^*\)) for each model, or include a small set of task-specific lengths that capture near-peak performance. Leaderboards could also incorporate a “context-efficiency” metric, rewarding models that achieve high accuracy with shorter sequences. This would make comparisons fairer across architectures with different context preferences (e.g., OPT vs. Mamba) and better reflect model capabilities across the diverse range of BabyLM evaluation tasks.

\section{Conclusion}

We present a systematic evaluation of sequence length sensitivity across BabyLM tasks, comparing the Transformer-based OPT and the state-space Mamba architectures. Our findings show that no single sequence length is universally optimal: shorter sequences often suffice for syntactic benchmarks like BLiMP, while longer contexts are necessary for tasks involving lexical productivity or discourse coherence. By identifying task-specific training-optimal lengths ($L^*$), we provide actionable guidance for balancing performance and efficiency in low-resource settings. Our results suggest that careful tuning of sequence length—rather than scaling alone—can yield meaningful gains in both compute and accuracy.

\section*{Limitation}

One limitation of our study is that we do not vary the mini-batch size or gradient accumulation strategy in conjunction with sequence length. While we vary sequence length to study its effect on task performance, it is possible to maintain a constant number of tokens per update by adjusting the mini-batch size or gradient accumulation steps. As a result, our experiments do not fully isolate the effect of sequence length from the effective batch size or the number of tokens processed per step.

Additionally, calibrating the optimal learning rate and schedule for each sequence length is challenging. Our experiments use a linear warmup proportional to sequence length, but we did not conduct exhaustive hyperparameter sweeps. It is possible that different learning rate or batch size configurations could change the relative performance of sequence lengths or architectures, and some of our reported training-optimal sequence lengths (\(L^{*}\)) may shift under alternative settings.

\section*{Acknowledgements}

Richard Diehl Martinez is supported by the Gates Cambridge Trust (grant OPP1144 from the Bill \& Melinda Gates Foundation). Suchir Salhan is supported by Cambridge University Press \& Assessment. Z\'ebulon Goriely is supported by an EPSRC DTP Studentship. This research was performed using resources provided by the Cambridge Service for Data Driven Discovery (CSD3) operated by the University of Cambridge Research Computing Service, provided by Dell EMC and Intel using Tier-2 funding from the Engineering and Physical Sciences Research Council (capital grant EP/T022159/1), and DiRAC funding from the Science and Technology Facilities Council.

\bibliography{custom}

\begin{thebibliography}{37}
\providecommand{\natexlab}[1]{#1}

\bibitem[{Armeni et~al.(2022)Armeni, Honey, and Linzen}]{armeni2022characterizing}
Kristijan Armeni, Christopher Honey, and Tal Linzen. 2022.
\newblock Characterizing verbatim short-term memory in neural language models.
\newblock In \emph{Proceedings of the 26th Conference on Computational Natural Language Learning (CoNLL)}, pages 405--424.

\bibitem[{Bengio et~al.(2009)Bengio, Louradour, Collobert, and Weston}]{bengio2009curriculum}
Yoshua Bengio, J{\'e}r{\^o}me Louradour, Ronan Collobert, and Jason Weston. 2009.
\newblock Curriculum learning.
\newblock In \emph{Proceedings of the 26th annual international conference on machine learning}, pages 41--48.

\bibitem[{Charpentier et~al.(2025)Charpentier, Choshen, Cotterell, Gul, Hu, Jumelet, Linzen, Liu, Mueller, Ross et~al.}]{charpentier2025babylm}
Lucas Charpentier, Leshem Choshen, Ryan Cotterell, Mustafa~Omer Gul, Michael Hu, Jaap Jumelet, Tal Linzen, Jing Liu, Aaron Mueller, Candace Ross, et~al. 2025.
\newblock Babylm turns 3: Call for papers for the 2025 babylm workshop.
\newblock \emph{arXiv preprint arXiv:2502.10645}.

\bibitem[{Chen et~al.(2021)Chen, Tsai, Bhojanapalli, Chung, Chang, and Ferng}]{chen2021simple}
Pu-Chin Chen, Henry Tsai, Srinadh Bhojanapalli, Hyung~Won Chung, Yin-Wen Chang, and Chun-Sung Ferng. 2021.
\newblock A simple and effective positional encoding for transformers.
\newblock \emph{arXiv preprint arXiv:2104.08698}.

\bibitem[{Chen et~al.(2023)Chen, Wong, Chen, and Tian}]{chen2023extending}
Shouyuan Chen, Sherman Wong, Liangjian Chen, and Yuandong Tian. 2023.
\newblock Extending context window of large language models via positional interpolation.
\newblock \emph{arXiv preprint arXiv:2306.15595}.

\bibitem[{Cheng et~al.(2023)Cheng, Aralikatte, Porada, Spinoso-Di~Piano, and Cheung}]{cheng2023mcbabylm}
Ziling Cheng, Rahul Aralikatte, Ian Porada, Cesare Spinoso-Di~Piano, and Jackie~CK Cheung. 2023.
\newblock \href {https://doi.org/10.18653/v1/2023.conll-babylm.18} {{M}c{G}ill {B}aby{LM} shared task submission: The effects of data formatting and structural biases}.
\newblock In \emph{Proceedings of the BabyLM Challenge at the 27th Conference on Computational Natural Language Learning}, pages 207--220, Singapore. Association for Computational Linguistics.

\bibitem[{Christiansen and Chater(2016)}]{christiansen2016now}
Morten~H Christiansen and Nick Chater. 2016.
\newblock The now-or-never bottleneck: A fundamental constraint on language.
\newblock \emph{Behavioral and brain sciences}, 39:e62.

\bibitem[{Clark et~al.(2025)Clark, Oh, and Schuler}]{clark2025linear}
Christian Clark, Byung-Doh Oh, and William Schuler. 2025.
\newblock Linear recency bias during training improves transformers’ fit to reading times.
\newblock In \emph{Proceedings of the 31st International Conference on Computational Linguistics}, pages 7735--7747.

\bibitem[{De~Varda and Marelli(2024)}]{de-varda-marelli-2024-locally}
Andrea De~Varda and Marco Marelli. 2024.
\newblock \href {https://doi.org/10.18653/v1/2024.cmcl-1.3} {Locally biased transformers better align with human reading times}.
\newblock In \emph{Proceedings of the Workshop on Cognitive Modeling and Computational Linguistics}, pages 30--36, Bangkok, Thailand. Association for Computational Linguistics.

\bibitem[{de~Varda et~al.(2024)de~Varda, Marelli, and Amenta}]{de2024cloze}
Andrea~Gregor de~Varda, Marco Marelli, and Simona Amenta. 2024.
\newblock Cloze probability, predictability ratings, and computational estimates for 205 english sentences, aligned with existing eeg and reading time data.
\newblock \emph{Behavior Research Methods}, 56(5):5190--5213.

\bibitem[{Devlin et~al.(2019)Devlin, Chang, Lee, and Toutanova}]{devlin-etal-2019-bert}
Jacob Devlin, Ming-Wei Chang, Kenton Lee, and Kristina Toutanova. 2019.
\newblock \href {https://doi.org/10.18653/v1/N19-1423} {{BERT}: Pre-training of deep bidirectional transformers for language understanding}.
\newblock In \emph{Proceedings of the 2019 Conference of the North {A}merican Chapter of the Association for Computational Linguistics: Human Language Technologies, Volume 1 (Long and Short Papers)}, pages 4171--4186, Minneapolis, Minnesota. Association for Computational Linguistics.

\bibitem[{Diehl~Martinez et~al.(2023)Diehl~Martinez, Goriely, McGovern, Davis, Caines, Buttery, and Beinborn}]{martinez-etal-2023-climb}
Richard Diehl~Martinez, Z{\'e}bulon Goriely, Hope McGovern, Christopher Davis, Andrew Caines, Paula Buttery, and Lisa Beinborn. 2023.
\newblock \href {https://doi.org/10.18653/v1/2023.conll-babylm.10} {{CLIMB} {--} curriculum learning for infant-inspired model building}.
\newblock In \emph{Proceedings of the BabyLM Challenge at the 27th Conference on Computational Natural Language Learning}, pages 112--127, Singapore. Association for Computational Linguistics.

\bibitem[{Elman(1990)}]{elman1990finding}
Jeffrey~L Elman. 1990.
\newblock Finding structure in time.
\newblock \emph{Cognitive Science}, 14(2):179--211.

\bibitem[{Geiping et~al.(2023)Geiping, Goldblum, Schwarzschild, Goldstein et~al.}]{geiping2023cramming}
Jonas Geiping, Micah Goldblum, Arjun Schwarzschild, Tom Goldstein, et~al. 2023.
\newblock Cramming: Training a language model on a single gpu in one day.
\newblock In \emph{Proceedings of the 40th International Conference on Machine Learning (ICML)}. PMLR.

\bibitem[{Gu and Dao(2024)}]{gumamba}
Albert Gu and Tri Dao. 2024.
\newblock Mamba: Linear-time sequence modeling with selective state spaces.
\newblock In \emph{First Conference on Language Modeling}.

\bibitem[{Haller et~al.(2024)Haller, Golde, and Akbik}]{haller-etal-2024-babyhgrn}
Patrick Haller, Jonas Golde, and Alan Akbik. 2024.
\newblock \href {https://aclanthology.org/2024.conll-babylm.7/} {{B}aby{HGRN}: Exploring {RNN}s for sample-efficient language modeling}.
\newblock In \emph{The 2nd BabyLM Challenge at the 28th Conference on Computational Natural Language Learning}, pages 82--94, Miami, FL, USA. Association for Computational Linguistics.

\bibitem[{Hao et~al.(2022)Hao, Sun, Dong, Han, Gu, and Wei}]{hao2022structured}
Yaru Hao, Yutao Sun, Li~Dong, Zhixiong Han, Yuxian Gu, and Furu Wei. 2022.
\newblock Structured prompting: Scaling in-context learning to 1,000 examples.
\newblock \emph{arXiv preprint arXiv:2212.06713}.

\bibitem[{Hendrycks et~al.(2020)Hendrycks, Burns, Basart, Zou, Mazeika, Song, and Steinhardt}]{hendrycks2020mmlu}
Dan Hendrycks, Christopher Burns, Steven Basart, Andy Zou, Mantas Mazeika, Dawn Song, and Jacob Steinhardt. 2020.
\newblock \href {https://arxiv.org/abs/2009.03300} {Measuring massive multitask language understanding}.
\newblock \emph{arXiv preprint arXiv:2009.03300}.

\bibitem[{Hofmann et~al.(2025)Hofmann, Weissweiler, Mortensen, Sch{\"u}tze, and Pierrehumbert}]{hofmann2025derivational}
Valentin Hofmann, Leonie Weissweiler, David~R Mortensen, Hinrich Sch{\"u}tze, and Janet~B Pierrehumbert. 2025.
\newblock Derivational morphology reveals analogical generalization in large language models.
\newblock \emph{Proceedings of the National Academy of Sciences}, 122(19):e2423232122.

\bibitem[{Izsak and Berend(2021)}]{izsak2021bertbudget}
Peter Izsak and G{\'a}bor Berend. 2021.
\newblock How to train bert with an academic budget.
\newblock In \emph{Proceedings of the 2021 Conference on Empirical Methods in Natural Language Processing (EMNLP)}.

\bibitem[{Kiros et~al.(2015)Kiros, Zhu, Salakhutdinov, Zemel, Urtasun, Torralba, and Fidler}]{kiros2015skipthought}
Ryan Kiros, Yukun Zhu, Ruslan Salakhutdinov, Richard Zemel, Raquel Urtasun, Antonio Torralba, and Sanja Fidler. 2015.
\newblock \href {https://papers.nips.cc/paper/5950-skip-thought-vectors.pdf} {Skip-thought vectors}.
\newblock In \emph{Advances in Neural Information Processing Systems}, volume~28.

\bibitem[{Li et~al.(2023)Li, Dong, Lin, and Guerin}]{li2023compressing}
Yucheng Li, Bo~Dong, Chenghua Lin, and Frank Guerin. 2023.
\newblock Compressing context to enhance inference efficiency of large language models.
\newblock \emph{arXiv preprint arXiv:2310.06201}.

\bibitem[{Mita et~al.(2025)Mita, Yoshida, and Oseki}]{mita2025developmentally}
Masato Mita, Ryo Yoshida, and Yohei Oseki. 2025.
\newblock Developmentally-plausible working memory shapes a critical period for language acquisition.
\newblock \emph{arXiv preprint arXiv:2502.04795}.

\bibitem[{Newport(1988)}]{newport1988constraints}
Elissa~L Newport. 1988.
\newblock Constraints on learning and their role in language acquisition: Studies of the acquisition of american sign language.
\newblock \emph{Language sciences}, 10(1):147--172.

\bibitem[{Press et~al.(2021)Press, Smith, and Lewis}]{press2021train}
Ofir Press, Noah~A Smith, and Mike Lewis. 2021.
\newblock Train short, test long: Attention with linear biases enables input length extrapolation.
\newblock \emph{arXiv preprint arXiv:2108.12409}.

\bibitem[{Salhan et~al.(2024)Salhan, Diehl~Martinez, Goriely, and Buttery}]{salhan-etal-2024-less}
Suchir Salhan, Richard Diehl~Martinez, Z{\'e}bulon Goriely, and Paula Buttery. 2024.
\newblock \href {https://aclanthology.org/2024.conll-babylm.15/} {Less is more: Pre-training cross-lingual small-scale language models with cognitively-plausible curriculum learning strategies}.
\newblock In \emph{The 2nd BabyLM Challenge at the 28th Conference on Computational Natural Language Learning}, pages 174--188, Miami, FL, USA. Association for Computational Linguistics.

\bibitem[{Samuel et~al.(2023)Samuel, Rekstad, and Velldal}]{samuel2023ltgbert}
David Samuel, Anders Rekstad, and Erik Velldal. 2023.
\newblock Trained on 100 million words and still in shape: Bert meets british national corpus (ltg-bert).
\newblock In \emph{Proceedings of the 17th Conference of the European Chapter of the Association for Computational Linguistics (EACL)}.

\bibitem[{Sennrich et~al.(2016)Sennrich, Haddow, and Birch}]{sennrich2016bpe}
Rico Sennrich, Barry Haddow, and Alexandra Birch. 2016.
\newblock \href {https://aclanthology.org/P16-1162} {Neural machine translation of rare words with subword units}.
\newblock In \emph{Proceedings of the 54th Annual Meeting of the Association for Computational Linguistics (ACL)}, pages 1715--1725. Association for Computational Linguistics.

\bibitem[{Soldaini et~al.(2024)Soldaini, Kinney, Bhagia, Schwenk, Atkinson, Authur, Bogin, Chandu, Dumas, Elazar, Hofmann, Jha, Kumar, Lucy, Lyu, Magnusson, Morrison, Muennighoff, Naik, Nam, Peters, Ravichander, Richardson, Shen, Strubell, Subramani, Tafjord, Walsh, Hajishirzi, Smith, Zettlemoyer, Beltagy, Groeneveld, Dodge, and Lo}]{soldaini2024dolma}
Luca Soldaini, Rodney Kinney, Akshita Bhagia, Dustin Schwenk, David Atkinson, Russell Authur, Ben Bogin, Khyathi Chandu, Jennifer Dumas, Yanai Elazar, Valentin Hofmann, Ananya~Harsh Jha, Sachin Kumar, Li~Lucy, Xinxi Lyu, Ian Magnusson, Jacob Morrison, Niklas Muennighoff, Aakanksha Naik, Crystal Nam, Matthew~E. Peters, Abhilasha Ravichander, Kyle Richardson, Zejiang Shen, Emma Strubell, Nishant Subramani, Oyvind Tafjord, Evan~Pete Walsh, Hannaneh Hajishirzi, Noah~A. Smith, Luke Zettlemoyer, Iz~Beltagy, Dirk Groeneveld, Jesse Dodge, and Kyle Lo. 2024.
\newblock {Dolma: An Open Corpus of Three Trillion Tokens for Language Model Pretraining Research}.
\newblock \emph{arXiv preprint}.

\bibitem[{Su et~al.(2024)Su, Ahmed, Lu, Pan, Bo, and Liu}]{su2024roformer}
Jianlin Su, Murtadha Ahmed, Yu~Lu, Shengfeng Pan, Wen Bo, and Yunfeng Liu. 2024.
\newblock Roformer: Enhanced transformer with rotary position embedding.
\newblock \emph{Neurocomputing}, 568:127063.

\bibitem[{Thamma and Heilbron(2025)}]{thamma2025human}
Abishek Thamma and Micha Heilbron. 2025.
\newblock Human-like fleeting memory improves language learning but impairs reading time prediction in transformer language models.
\newblock \emph{arXiv preprint arXiv:2508.05803}.

\bibitem[{Thoma et~al.(2023)Thoma, Weyers, {\c{C}}ano, Schweter, Mueller, and Roth}]{thoma-etal-2023-cogmemlm}
Lukas Thoma, Ivonne Weyers, Erion {\c{C}}ano, Stefan Schweter, Jutta~L Mueller, and Benjamin Roth. 2023.
\newblock \href {https://doi.org/10.18653/v1/2023.conll-babylm.15} {{C}og{M}em{LM}: Human-like memory mechanisms improve performance and cognitive plausibility of {LLM}s}.
\newblock In \emph{Proceedings of the BabyLM Challenge at the 27th Conference on Computational Natural Language Learning}, pages 180--185, Singapore. Association for Computational Linguistics.

\bibitem[{Warstadt et~al.(2023)Warstadt, Mishra, Yoshida, Gauthier, and et~al.}]{warstadt2023babylm}
Alex Warstadt, Sweta~Agrawal Mishra, Masato Yoshida, Jon Gauthier, and et~al. 2023.
\newblock Findings of the babylm challenge: Sample-efficient pretraining on developmentally plausible corpora.
\newblock In \emph{Proceedings of the 2023 Conference on Empirical Methods in Natural Language Processing (EMNLP)}.

\bibitem[{Warstadt et~al.(2020)Warstadt, Parrish, Liu, Mohananey, Peng, Wang, and Bowman}]{warstadt-etal-2020-blimp-benchmark}
Alex Warstadt, Alicia Parrish, Haokun Liu, Anhad Mohananey, Wei Peng, Sheng-Fu Wang, and Samuel~R. Bowman. 2020.
\newblock \href {https://doi.org/10.1162/tacl_a_00321} {{BL}i{MP}: The benchmark of linguistic minimal pairs for {E}nglish}.
\newblock \emph{Transactions of the Association for Computational Linguistics}, 8:377--392.

\bibitem[{Wilcox et~al.(2025)Wilcox, Hu, Mueller, Warstadt, Choshen, Zhuang, Williams, Cotterell, and Linzen}]{wilcox2025bigger}
Ethan~Gotlieb Wilcox, Michael~Y Hu, Aaron Mueller, Alex Warstadt, Leshem Choshen, Chengxu Zhuang, Adina Williams, Ryan Cotterell, and Tal Linzen. 2025.
\newblock Bigger is not always better: The importance of human-scale language modeling for psycholinguistics.
\newblock \emph{Journal of Memory and Language}, 144:104650.

\bibitem[{Zellers et~al.(2019)Zellers, Holtzman, Bisk, Farhadi, and Choi}]{zellers2019hellaswag}
Rowan Zellers, Ari Holtzman, Yonatan Bisk, Ali Farhadi, and Yejin Choi. 2019.
\newblock \href {https://doi.org/10.18653/v1/P19-1472} {Hellaswag: Can a machine really finish your sentence?}
\newblock In \emph{Proceedings of the 57th Annual Meeting of the Association for Computational Linguistics}, pages 4791--4800.

\bibitem[{Zhang et~al.(2022)Zhang, Roller, Goyal, Artetxe, Chen, Chen, Dewan, Ghazvininejad, Guti{\'e}rrez, Hazard et~al.}]{zhang2022opt}
Susan Zhang, Stephen Roller, Naman Goyal, Mikel Artetxe, Moya Chen, Shuohui Chen, Shu Dewan, Marjan Ghazvininejad, Sinong Guti{\'e}rrez, Lucy Hazard, et~al. 2022.
\newblock \href {https://arxiv.org/abs/2205.01068} {Opt: Open pre-trained transformer language models}.
\newblock \emph{arXiv preprint arXiv:2205.01068}.

\end{thebibliography}
\newpage 
\appendix
\onecolumn  

\newpage 

\section{Training Setup: Hyperparameters}

\begin{table*}[ht]
\centering
\renewcommand{\arraystretch}{0.9}
\caption{Training hyperparameters for BabyLM experiments. This table summarizes model, training, checkpointing, hardware, and dataset settings.}
\label{tab:hyperparams_main}
\begin{tabular}{llp{0.4\textwidth}}
\hline
\textbf{Category} & \textbf{Parameter} & \textbf{Value / Notes} \\
\hline
\multirow{4}{*}{Model} 
& Type & 
\begin{itemize}[leftmargin=*,noitemsep,topsep=0pt]
\item OPT: 12-layer, 768 hidden, 12 heads, FFN 3072
\item Mamba: 32-layer, 768 hidden
\end{itemize} \\
& Vocabulary size & 50,257 tokens \\
& Max sequence length & 64--16,384 tokens, varies per experiment \\
& Pretrained weights & Random initialization \\
\hline
\multirow{7}{*}{Training} 
& Epochs & 10 \\
& Global batch size & 64 sequences \\
& Per-device batch size & $\frac{\text{GLOBAL\_BATCH\_SIZE}} {(\text{num\_devices} \times \text{accumulation\_steps})}$ \\
& Gradient accumulation steps & 1 (configurable via CLI) \\
& Learning rate & Scales linearly with seq. length if warmup: $5\times10^{-5} \times \frac{\text{seq\_len}}{64}$ \\
& Tokens per batch & $\text{GLOBAL\_BATCH\_SIZE} \times \text{seq\_len}$ \\
& Tokens per update & Tokens per batch × accumulation steps \\
\hline
\multirow{3}{*}{Checkpointing} 
& Frequency & Every 1M, 10M, 100M tokens (CustomCheckpointingCallback) \\
& Hub push & Optional via CLI \\
& Resume from checkpoint & Supported \\
\hline
\multirow{3}{*}{Hardware / Precision} 
& Devices & 4 (configurable via CLI) \\
& Mixed precision & bf16 (DeepSpeed / Trainer) \\
& DeepSpeed & Optional, stage 3 ZeRO with CPU offload \\
\hline
\multirow{3}{*}{Dataset} 
& Source & Hugging Face pretokenized datasets \\
& Examples & \href{https://huggingface.co/babylm-seqlen/}{babylm-seqlen/} \\
&  & \href{https://huggingface.co/babylm-seqlen/}{train\_100M\_<seq\_len>\_single\_shuffle} \\

& Preprocessing & Labels set as input\_ids for causal LM training \\
\hline
\end{tabular}
\end{table*}

\newpage 
\section{Dataset Statistics} \label{dataset-statistics}

\begin{table}[h!]
\centering
\begin{tabular}{lc}
\toprule
Sequence Length & Num Sequences \\
\midrule
\href{https://huggingface.co/datasets/babylm-seqlen/train_100M_64_single_shuffle}{64} & 2,556,406 \\
\href{https://huggingface.co/datasets/babylm-seqlen/train_100M_128_single_shuffle}{128} & 1,278,130 \\
\href{https://huggingface.co/datasets/babylm-seqlen/train_100M_256_single_shuffle}{256} & 639,002 \\
\href{https://huggingface.co/datasets/babylm-seqlen/train_100M_512_single_shuffle}{512} & 319,435 \\
\href{https://huggingface.co/datasets/babylm-seqlen/train_100M_1024_single_shuffle}{1024} & 159,656 \\
\href{https://huggingface.co/datasets/babylm-seqlen/train_100M_2048_single_shuffle}{2048} & 79,761 \\
\href{https://huggingface.co/datasets/babylm-seqlen/train_100M_4096_single_shuffle}{4096} & 39,814 \\
\href{https://huggingface.co/datasets/babylm-seqlen/train_100M_8192_single_shuffle}{8192} & 19,844 \\
\href{https://huggingface.co/datasets/babylm-seqlen/train_100M_16384_single_shuffle}{16384} & 9,863 \\
\bottomrule
\end{tabular}
\caption{Number of sequences for each fixed sequence length dataset. Sequence lengths are clickable links to the corresponding Hugging Face dataset.}
\end{table}

\begin{table}[ht]
\centering
\renewcommand{\arraystretch}{0.9}
\caption{Example settings for per-device batch size, learning rate, and tokens per batch at different sequence lengths.}
\label{tab:hyperparams_examples}
\begin{tabular}{cccc}
\hline
\textbf{Seq Length} & \textbf{Per-Device Batch} & \textbf{Learning Rate} & \textbf{Tokens per Batch} \\
\hline
64 & 16 & 5e-5 & 4,096 \\
128 & 16 & 1e-4 & 8,192 \\
512 & 16 & 4e-4 & 32,768 \\
2048 & 16 & 1.6e-3 & 131,072 \\
8192 & 16 & 6.4e-3 & 524,288 \\
\hline
\end{tabular}
\end{table}

\newpage 
\section{Final Checkpoint Results: OPT and Mamba (\(\pm\) Warmup)}

\textit{Table} \ref{tab:mamba_opt_results_split} provides a detailed breakdown of model performance on the full zero-shot evaluation tasks. In particular, we report differences between training models with and without warmup.  

\begin{table*}[htbp]
\centering
\small
\begin{tabular}{|c|c|r|r|r|r|r|r|}
\hline
\textbf{Model} & \textbf{Warmup} & \textbf{Seq Len} & \textbf{BLiMP} & \textbf{BLiMP Suppl.} & \textbf{Entity Tracking} & \textbf{EWoK} & \textbf{Wug} \\
\hline
mamba & + & 64 & 68.33 & 63.20 & 19.05 & 50.66 & 63.50 \\ 
mamba & - & 64 & 69.56 & 61.23 & 22.24 & 51.05 & 62.00 \\
mamba & + & 128 & 67.31 & 60.40 & 23.20 & 50.1 & 54.00 \\ 
mamba & - & 128 & 69.87 & 57.94 & 38.69 & 51.34 & 70.50 \\
mamba & + & 256 & 69.19 & 61.60 & 12.48 & 51.34 & 56.50 \\ 
mamba & - & 256 & 69.14 & 60.04 & 25.28 & 51.28 & 53.50 \\
mamba & + & 512 & 68.87 & 59.60 & 16.33 & 52.54 & 51.50 \\ 
mamba & - & 512 & 68.45 & 60.98 & 23.16 & 49.99 & 55.50 \\
mamba & + & 1024 & 67.30 & 58.00 & 31.95 & 52.31 & 57.50 \\  
mamba & - & 1024 & 66.28 & 56.99 & 21.52 & 50.35 & 62.50 \\
mamba & + & 2048 & 71.62 & 60.40 & 14.07 & 51.82 & 53.50 \\  
mamba & - & 2048 & 63.33 & 55.03 & 20.25 & 50.30 & 56.50 \\
mamba & + & 4096 & 69.56 & 57.60 & 13.93 & 51.49 & 49.00 \\ 
mamba & - & 4096 & 59.10 & 55.50 & 17.80  & 50.18 & 62.00  \\
mamba & + & 8192 & 66.91 & 59.20 & 22.70 & 51.05 & 51.50 \\ 
mamba & - & 8192 & 59.21 & 52.94 & 23.37 & 49.83 & 61.50 \\
\hline
opt & + & 64 & 70.21 & 67.60 & -- & 51.82 & 48.00 \\ 
opt & - & 64 & 75.44 & 66.45 & -- & 51.64 & 49.50 \\
opt & + & 128 & 70.78 & 66.80 & -- & 51.92 & 46.50 \\ 
opt & - & 128 & 74.87 & 63.53 & -- & 51.98 & 45.00 \\
opt & + & 256 & 73.88 & 67.20 & 32.42 & 52.18 & 44.50 \\ 
opt & - & 256 & 73.11 & 59.92 & 20.93 & 51.68 & 46.00 \\
opt & + & 512 & 71.9 & 59.60 & 26.80 & 51.45 & 47.50 \\ 
opt & - & 512 & 70.63 & 61.70 & 26.99 & 51.80 & 47.00 \\
opt & + & 1024 & 72.69 & 62.40 & 26.15 & 51.28 & 48.5 \\ 
opt & - & 1024 & 68.23 & 57.79 & 26.27 & 50.66 & 50.00 \\
opt & + & 2048 & 72.05 & 62.40 & 25.96 & 52.37 & 45.50 \\ 
opt & - & 2048 & 61.67 & 57.23 & 29.57 & 49.89 & 50.50 \\
opt & + & 4096 & 56.25 & 48.0 & 40.23 & 49.70 & 90.00 \\ 
opt & - & 4096 & 58.58 & 54.58 & 17.03 & 50.10 & 66.00 \\
opt & + & 8192 & 55.05 & 50.40 & 40.38 & 50.89 & 66.00 \\ 
opt & - & 8192 & 56.01 & 53.21 & 19.38 & 49.70 & 64.50 \\
\hline
\end{tabular}
\caption{Evaluation results across multiple benchmarks for Mamba and OPT models. ‘--’ denotes missing data (NaN).}
\label{tab:mamba_opt_results_split}
\end{table*}

\newpage 
\section{Learning Dynamics: Task Evaluation on Checkpoints }

\begin{figure}[!ht]
    \caption{Comparison of the performance of Mamba and OPT models on BabyLM Evaluation tasks throughout training. Checkpoints are saved at increasingly intervals throughout training: every 1M words until 10M words are seen, every 10M words until 100M words are seen, and every 100M words until 1B words are seen.  }
    \centering
    \includegraphics[width=0.85\linewidth]{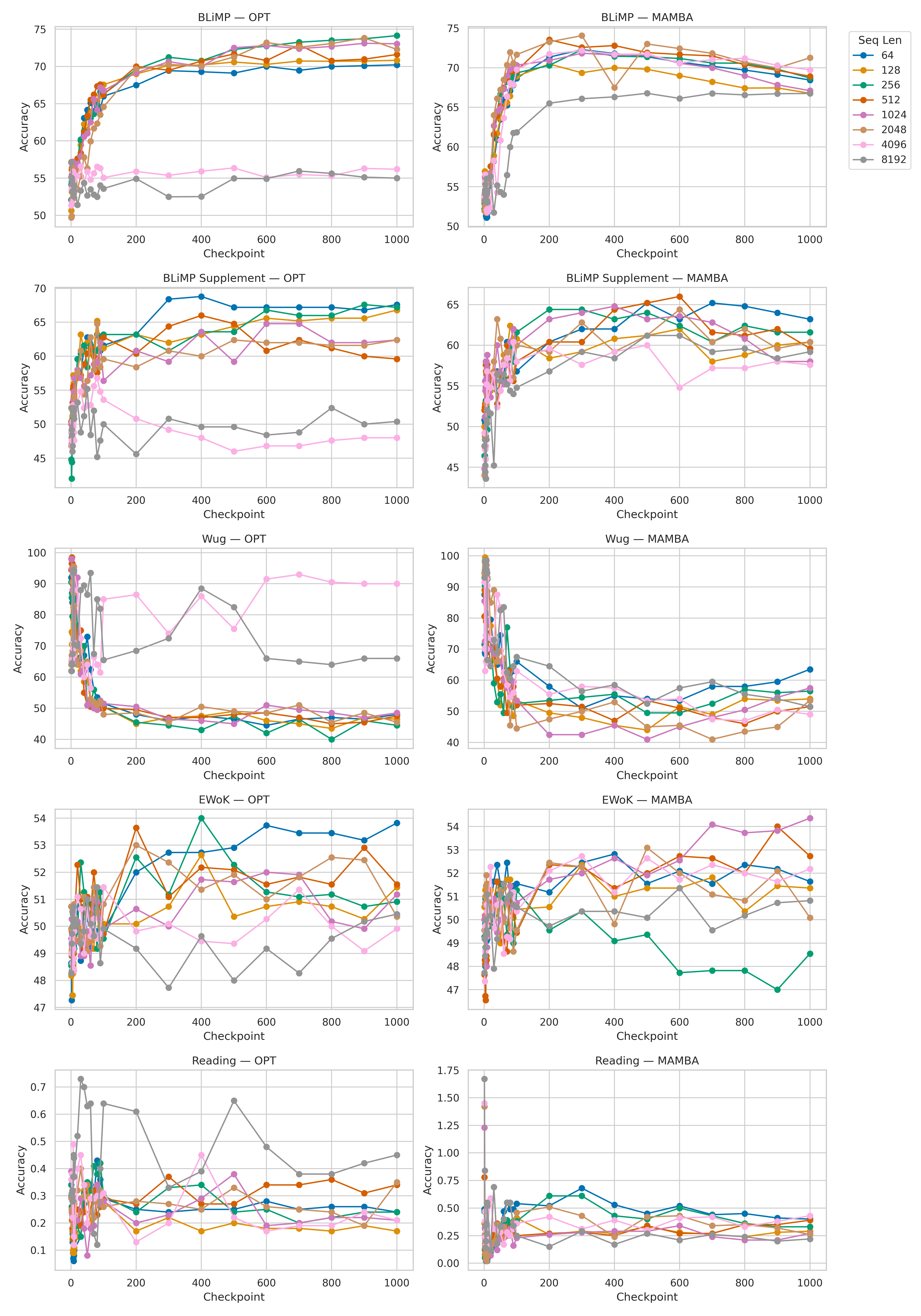}
    
    \label{fig:checkpoints-all}
\end{figure}

\newpage

\section{Subtask Accuracy for OPT and Mamba Families}

\begin{figure*}[!ht]
    \centering
    \includegraphics[width=0.95\linewidth]{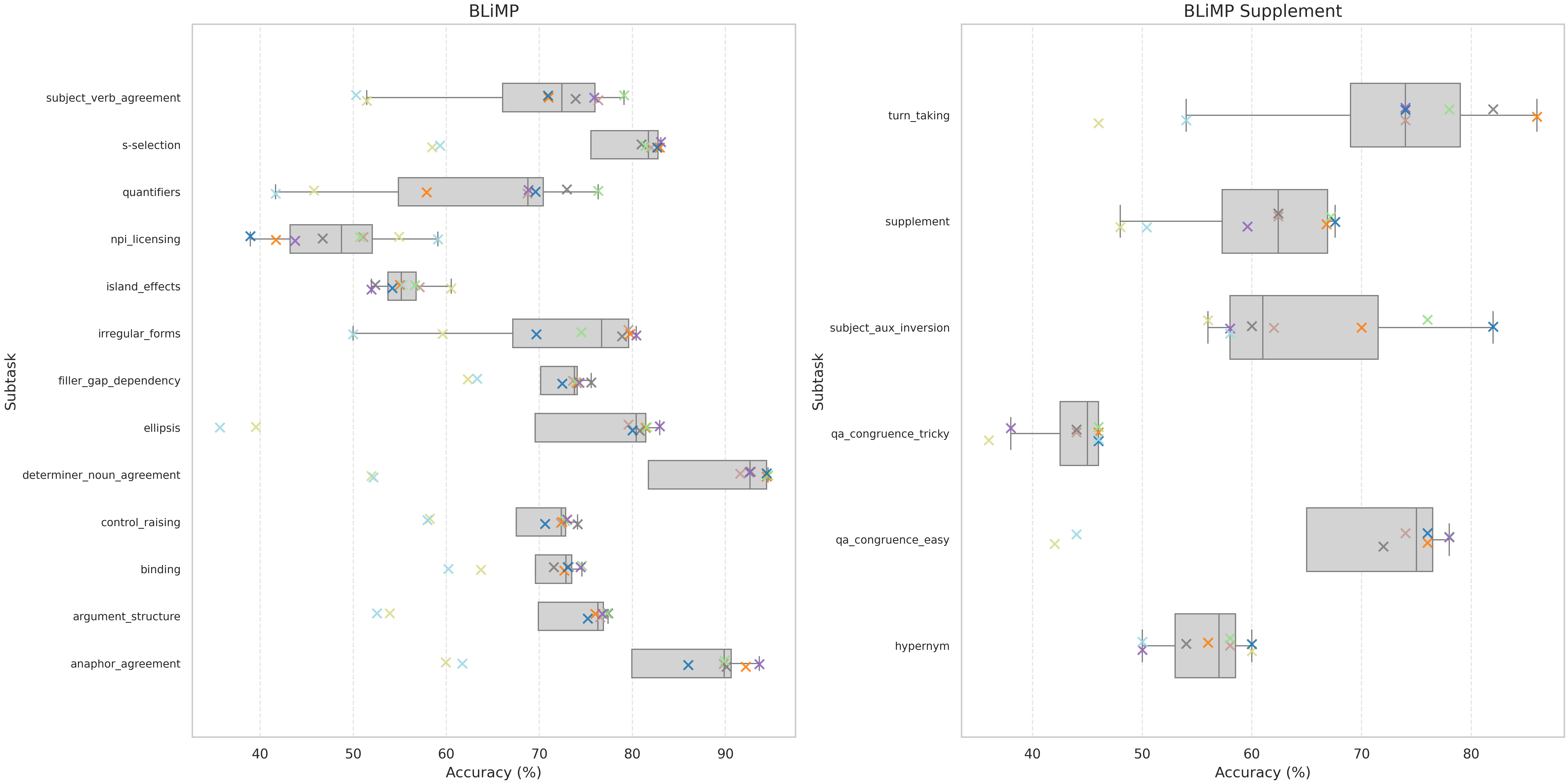}
    \caption{Distribution of OPT Sequence Length Model Accuracies on BLiMP and BLiMP Supplement}
    \label{fig:checkpoints-all-1}
\end{figure*}

\begin{figure*}[!ht]
    \centering
    \includegraphics[width=0.95\linewidth]{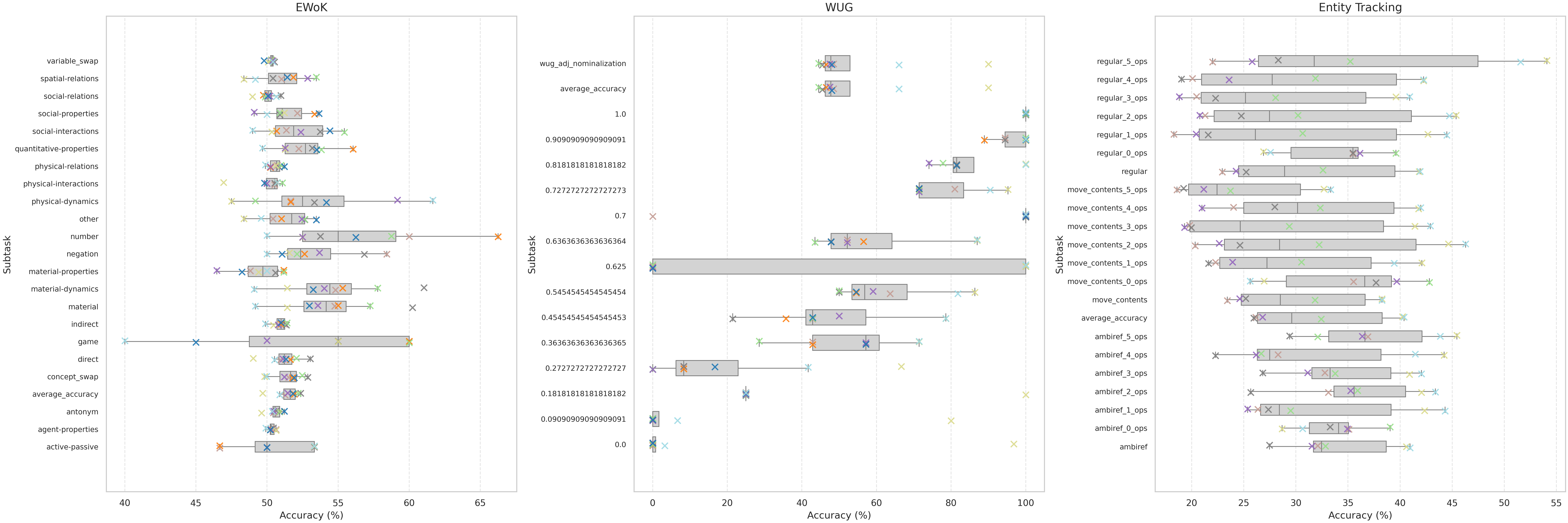}
    \caption{Distribution of EWoK, Wug and Entity Tracking Sequence Length Model Accuracies for OPT Architecture}
    \label{fig:checkpoints-all-2}
\end{figure*}

\newpage 

\section{F1 Scores for Fine-Tuning}

\begin{figure}[!ht]
    \centering
        \caption{\textbf{F1 for Fine-Tuned Models}}
        \label{fig:finetune-f1}
        \centering
        \includegraphics[width=\linewidth]{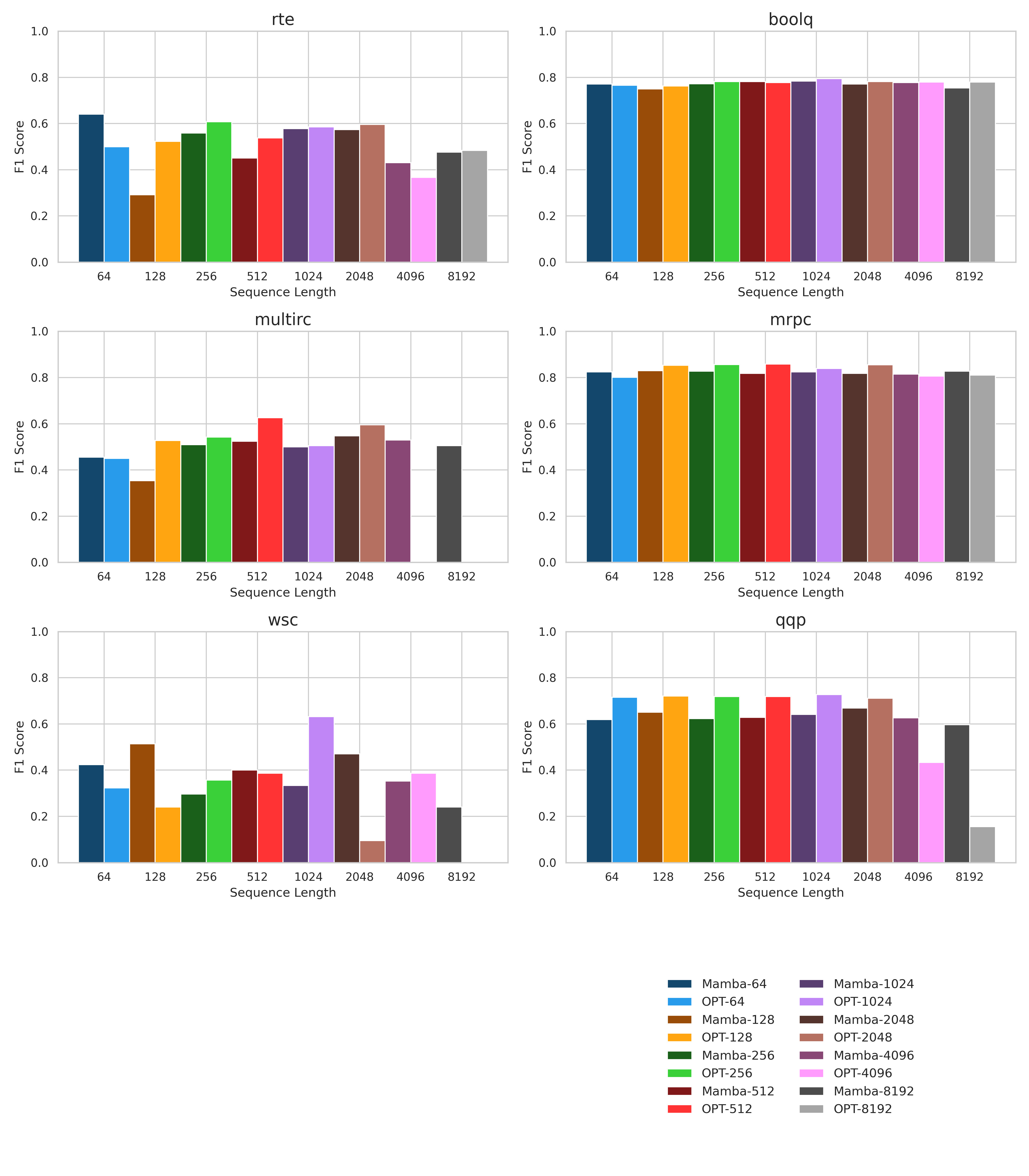}
        \caption{F1 Scores for OPT and Mamba Families on Fine-Tuned Tasks}
\end{figure}

\end{document}